\documentclass[conference]{IEEEtran}
\usepackage{cite}

\ifCLASSINFOpdf
\else
\fi
\usepackage{amsmath}
\usepackage{algorithmic}

\usepackage{array}

\usepackage[export]{adjustbox}
\usepackage{xcolor}
\usepackage{minibox}
\usepackage{booktabs}

\usepackage[font=small,skip=0pt]{caption}
\usepackage{url}

\begin{document}
\bstctlcite{IEEEexample:BSTcontrol}
%
\title{Judging a Book by its Cover}

  
\author{
   \IEEEauthorblockN{
     Brian Kenji Iwana\IEEEauthorrefmark{1}, 
     Syed Tahseen Raza Rizvi\IEEEauthorrefmark{3}\IEEEauthorrefmark{4},
     Sheraz Ahmed\IEEEauthorrefmark{3},
     Andreas Dengel\IEEEauthorrefmark{3}\IEEEauthorrefmark{4},
     Seiichi Uchida\IEEEauthorrefmark{1}
   }
   \IEEEauthorblockA{
     \IEEEauthorrefmark{1}Department of Advanced Information Technology, Kyushu University, Fukuoka, Japan\\ Email: \{brian, uchida\}@human.ait.kyushu-u.ac.jp
   }
   \IEEEauthorblockA{
     \IEEEauthorrefmark{3}German Research Center for Artificial Intelligence, Kaiserlautern, Germany\\ Email: \{syed\_tahseen\_raza.rizvi, Sheraz.Ahmed, Andreas.Dengel\}@dfki.de
   }
   \IEEEauthorblockA{
     \IEEEauthorrefmark{4}Kaiserslautern University of Technology, Kaiserlautern, Germany
   }
 }


%


\maketitle

\begin{abstract}
Book covers communicate information to potential readers, but can that same information be learned by computers? We propose using a deep Convolutional Neural Network (CNN) to predict the genre of a book based on the visual clues provided by its cover. The purpose of this research is to investigate whether relationships between books and their covers can be learned. However, determining the genre of a book is a difficult task because covers can be ambiguous and genres can be overarching. Despite this, we show that a CNN can extract features and learn underlying design rules set by the designer to define a genre. Using machine learning, we can bring the large amount of resources available to the book cover design process. In addition, we present a new challenging dataset that can be used for many pattern recognition tasks.
\end{abstract}


%
\IEEEpeerreviewmaketitle

\section{Introduction}
``Don't judge a book by its cover'' is a common English idiom meaning not to judge something by its outward appearance.
Although, it still happens when a reader encounters a book.
The cover of a book is often the first interaction and it creates an impression on the reader.
It starts a conversation with a potential reader and begins to draw a story revealing the contents within.
But, what does the book cover say?
What are the clues that the book cover reveals?
While the visual clues can communicate information to humans, we explore the possibility of using computers to learn about a book by its cover.
Machine learning provides the ability to use a large amount of resources to the world of design.
By bridging the gap between design and machine learning, we hope to use a large dataset to understand the secrets of visual design.

We propose a method deriving a relationship between book covers and their genre automatically.
The goal is to determine if genre information can be learned based on the visual aspects of a cover created by the designer. 
This research can aid the design process by revealing underlying information, help promotion and sales processes by providing automatic genre suggestion, and be used in computer vision fields.

The difficulty of this task is that books come with a wide variety of book covers and styles, including nondescript and misleading covers.
Unlike other object detection and classification tasks, genres are not concretely defined.
Another problem is that there is a massive amount of books exist and it is not suitable for exhaustive search methods.

To tackle this task, we present the use of an artificial neural network.
The concept of neural networks and neural coding is to use interconnected nodes to work together to capture information.
Early neural network-like models such as multilayer perception learning were invented in the 1970s but fell out of favor~\cite{schmidhuber2015deep}.
More recently, artificial neural networks have been a focus of state-of-the-art research because of their successes in pattern recognition and machine learning.
Their successes are in part due to the increase in data availability, increase in processing power, and introduction of GPUs~\cite{chellapilla2006high}.

Convolutional Neural Networks (CNN)~\cite{lecun1998gradient}, in particular, are multilayer neural networks that utilize learned convolutional kernels, or filters, as a method of feature extraction.
The general idea is to use learned features rather than pre-designed features as the feature representation for image recognition.
Recent deep CNNs combine multiple convolutional layers along with fully-connected layers.
By increasing the depth of the network, higher level features can be learned and discriminative parts of the images are exaggerated~\cite{zeiler2014visualizing}.
These deep CNNs have had successes in many fields including digit-recognition~\cite{lecun1998gradient,ciresan2012multi} and large-scale image recognition~\cite{szegedy2015going,simonyan2014very}.

The contribution of this paper is to demonstrate that connections between book genres can be learned using only the cover images.
To solve this task, we used the concept of transfer learning and developed a CNN based system for book cover genre classification.
AlexNet~\cite{krizhevsky2012imagenet} pre-trained on ImageNet~\cite{imagenet_cvpr09} is adapted for the task of genre recognition.
We also reveal the relationships automatically learned between genres and book covers.

Secondly, we created a large dataset containing 137,788 books in 32 classes made of book cover images, title text, author text, and category membership. 
This dataset is very challenging and can be used for a variety of tasks some of which include text recognition, font analysis, and genre prediction.
Furthermore, although AlexNet pre-trained on ImageNet has already achieved state-of-the-art results on document classification~\cite{afzal2015deepdocclassifier,kang2014convolutional}, we had a limited accuracy which indicates the high level of difficulty of the proposed dataset.

The remaining of this paper is organized as follows. 
Section~\ref{related} provides related works in design learning with machine learning.
Section~\ref{cnn} elaborates on CNNs and the details of the proposed method.
In Section~\ref{results}, we confirm the proposed method and analyzed the experimental results.
The book cover designed principles learned by the CNN is detailed in Section~\ref{design}.
Finally, Section~\ref{conclusion} draws the conclusion.

\section{Related Works}
\label{related}

Visual design is intentional and serves a purpose.
It has a rich history and exploring the purposes of design has been extensively analyzed by designers~\cite{bookhistory} but is a relatively new field in machine learning.

Techniques have been used to identify artistic styles and qualities of paintings and photographs~\cite{karayev2013recognizing,gatys2015neural,datta2006studying,datta2008image}.
Gatys, et al.~\cite{gatys2015neural} used deep CNNs to learn and copy the artistic style of paintings.
Similarly, the goal of this trial is to learn the stylistic qualities of the work, but we go beyond to learn the underlying meaning behind the style.

In the field of genre classification, there have been attempts to classify music by genre~\cite{tzanetakis2002musical,mckay2004automatic,pye2000content}.
It was also done in the fields of paintings~\cite{zujovic2009classifying,karayev2013recognizing} and text~\cite{finn2006learning,petrenz2011stable}.
However, most of these methods use designed features or features specific to the task.
In a more general sense, document classification tackles a similar problem in that it classes documents into architectural categories.
In particular, deep CNNs have been successful in document classification~\cite{afzal2015deepdocclassifier,kang2014convolutional}.
Harley et al.~\cite{harley2015evaluation}, used a region-based CNNs to guide the document classification.

\section{Convolutional Neural Networks}
\label{cnn}
Modern CNNs are made up of three components: convolutional layers, pooling layers, and fully-connected layers. The convolutional layers consist of feature maps produced by repeatedly applying filters across the input. The filters represent shared weights and are trained using backpropagation. 
The feature maps resulting from the applied filters are down-sampled by a max pooling layer to reduce redundancy improving the computational time for future layers. 
Finally, the last few layers of a CNN are made up of fully-connected layers. 
These layers are given a vector representation of the images from a preceding pooling layer and continue like standard feedforward neural networks.

\subsection{AlexNet}
The network used for our book cover classification is inspired from the work of Krizhevsky et al.~\cite{krizhevsky2012imagenet} 
We used a pre-trained network on ImageNet~\cite{imagenet_cvpr09}.
By pre-training AlexNet on a very large dataset such as ImageNet, its possible to take advantage of the learned features and transfer it to other applications.
Initializing a network with transferred features has shown to improve generalization~\cite{yosinski2014transferable}.
To accomplish this, we remove the original softmax output layer for the 1,000-class classification of ImageNet and replace it with a 30-class softmax for the experiment.
Subsequently, the training is continued using the pre-trained parameters as an initialization.

The network architecture is as follows.
The network consists of a total of eight layers, where the first five are convolutional layers followed by three fully-connected layers. 
Of the five convolutional layers, the first and second layers are made of 96 filters of size $11 \times 11 \times 3$ stride 4, and $5 \times 5 \times 48$ stride 1 respectively and are response-normalized.
The last three convolutional layers have 384, 384, and 256 nodes and use filters of size $3 \times 3 \times 192$.
These last three convolutional layers do not use any normalization or pooling.
The final three fully-connected layers have 4,096 nodes each.
Both the convolutional layers and the fully-connected layers have Rectified Linear Unit (ReLU) activation functions.
Dropout with a keep probability of 0.5 is used for the first two fully-connected layers.

The model was trained with gradient decent with an initial learning rate of 0.01, after which, the learning rate was divided by 10 every 100,000 iterations.
The reported results were taken after 450,000 iterations.
Also, a weight decay of 0.0005 and momentum of 0.9 was implemented.
The update rule for each weight $w$ is defined as~\cite{krizhevsky2012imagenet}:
\begin{align}
v_{i+1} &= 0.9 v_i-0.0005 \epsilon w_i - \epsilon \left<\frac{\partial L}{\partial w}\bigr|_{w_i}\right> \\
w_{i+1} &= w_i + v_{i+1}.
\end{align}

\subsection{LeNet}
For a comparison, we trained a network similar to a LeNet~\cite{lecun1998gradient}.
This CNN used input images, that were scaled to 56px by 56px, in batches of 200.
There were three convolutional layers with 32 nodes, 64 nodes, and 128 nodes respectively.
Each convolutional layer uses a filter size of $5 \times 5 \times 1$ at stride 1 and were proceeded by maxpooling layers of $2 \times 2$ stride 1.
The network concluded with a 1024 node full-convolutional layer and a softmax output layer.
Each layer used ReLU activations and a constant learning rate of 0.0001.
Dropout with a keep probability of 0.5 was used after the fully-connected layer.
Finally, the network was trained for 30,000 iterations of using an Adam optimizer~\cite{kingma2014adam}. 
The modified LeNet was trained on the same training set and tested with the same test set as the AlexNet experiment.

\section{Experimental results}
\label{results}

\subsection{Dataset preparation}
\label{results:prep}
The dataset was collected from the book cover images and genres listed by Amazon.com~\cite{amazon}.
The full dataset contains 137,788 unique book cover images in 32 classes as well as the title, author, and subcategories for each respective book.
Each book's class is defined as the top categories under ``Books'' in the Amazon.com marketplace.
However, for the experiment we refined the dataset into 30 classes of 1,900 books in each class. 
The 30 classes, or genres, used in the experiment are listed in Table~\ref{genacc}.
To equalize the number of books in each class, books were chosen at random to be included in the experiment. 
The two categories, ``Gay \& Lesbian'' and ``Education \& Teaching,'' were not used for the experiment because they only contain 1,341 and 1,664 books respectively, thus not having enough representation in the dataset. 

\begin{table}
\caption{Top 1 Genre Accuracy Comparison}
\label{genacc}
\centering
\begin{tabular}{lccccc}
\hline
& \multicolumn{2}{c}{\textbf{LeNet}} & & \multicolumn{2}{c}{\textbf{AlexNet}} \\
\cmidrule{2-3} 
\cmidrule{5-6} 
\textbf{Genre} & \textbf{Top 1} & \textbf{Top 3} & & \textbf{Top 1} & \textbf{Top 3} \\
\hline
Arts \& Photography&5.8 & 11.6 &&12.1&31.1\\
Biographies \& Memoirs&5.3&18.4 &&13.2&29.5\\
Business \& Money&10.0&25.3&&12.6&25.8\\
Calendars&18.9&37.9&&47.9&65.3\\
Children's Books&24.7&42.1&&42.1&61.6\\
Comics \& Graphic Novels&15.8&33.7&&47.4&67.9\\
Computers \& Technology&29.5&42.8&&44.7&59.5\\
Cookbooks, Food \& Wine&14.2&32.6&&43.7&57.4\\
Crafts, Hobbies \& Home&7.4&22.1&&17.4&36.8\\
Christian Books \& Bibles&8.4&23.7&&7.4&26.3\\
Engineering \& Transportation&10.0&21.1&&20.0&34.7\\
Health, Fitness \& Dieting&4.2&15.8&&12.6&29.5\\
History&6.3&16.8&&12.6&27.9\\
Humor \& Entertainment&5.3&16.3&&10.5&22.6\\
Law&14.7&25.8&&25.3&38.4\\
Literature \& Fiction&3.2&12.1&&11.1&22.6\\
Medical Books&12.6&30.0&&19.5&36.8\\
Mystery, Thriller \& Suspense&23.7&40.0&&34.2&48.9\\
Parenting \& Relationships&14.7&35.3&&24.2&39.5\\
Politics \& Social Sciences&3.7&18.4&&6.8&21.6\\
Reference&13.2&26.8&&20.0&34.2\\
Religion \& Spirituality&8.4&27.9&&16.3&31.6\\
Romance&27.4&43.2&&45.3&60.5\\
Science \& Math&8.4&26.3&&14.2&29.5\\
Science Fiction \& Fantasy&14.7&33.2&&35.8&52.6\\
Self-Help&13.7&31.6&&14.2&33.2\\
Sports \& Outdoors&5.3&16.8&&14.7&28.4\\
Teen \& Young Adult&7.9&17.4&&12.1&28.4\\
Test Preparation&47.9&56.8&&68.9&78.4\\
Travel&19.5&33.7&&33.2&48.4\\
\textbf{Total Average} & \textbf{13.5}&\textbf{27.8}&& \textbf{24.7}&\textbf{40.3} \\
\hline
\end{tabular}
\end{table}


Also, when the dataset was collected, each book was assigned to only a single category. 
If the book belonged to multiple categories, one was chosen at random.
We randomized and split the dataset into 90\% training set and 10\% test set.
No pruning of cover images and no class membership corrections were done.
In addition, we resized all of the images to fit 227px by 227px by 3 color channels for the input of the AlexNet and 56px by 56px by 3 color channels for LeNet.

\subsection{Evaluation}


The pre-trained AlexNet with transfer learning resulted in a test set Top~1 classification accuracy of 24.7\% , 33.1\% for Top~2, and 40.3\% for Top~3 which are 7.4, 5.0, and 4.0 times better than random chance respectively.
As comparison, using the modified LeNet, we had a Top~1 accuracy of 13.5\%, Top~2 accuracy of 21.4\%, and Top~3 accuracy of 27.8\%.
The AlexNet performed much better on this dataset than the LeNet.
Considering that CNN solutions are state-of-the-art for image and document recognition, the results show that classification of book cover designs is possible, although a very difficult task.

Table~\ref{genacc} shows the individual Top~1 accuracies for each genre.
In every class except ``Christian Books \& Bibles,'' the AlexNet performed better.
For most cases, AlexNet had more than twice as good Top~1 accuracy compared to LeNet.

\subsection{Analysis}

\begin{figure}
\begin{center}
\setlength\fboxsep{0pt}
\setlength\fboxrule{0pt}
\framebox[0.15\columnwidth]{\includegraphics[width=0.15\columnwidth]{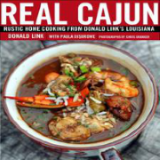}}
\framebox[0.15\columnwidth]{\includegraphics[width=0.15\columnwidth]{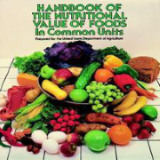}}
\framebox[0.15\columnwidth]{\includegraphics[width=0.15\columnwidth]{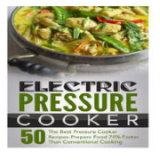}}
\framebox[0.024\columnwidth]{ }
\framebox[0.15\columnwidth]{\includegraphics[width=0.15\columnwidth]{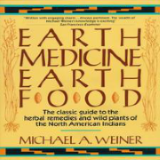}}
\framebox[0.15\columnwidth]{\includegraphics[width=0.15\columnwidth]{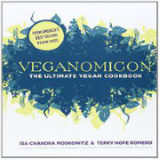}}
\framebox[0.15\columnwidth]{\includegraphics[width=0.15\columnwidth]{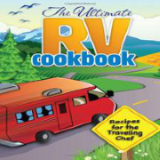}}
\end{center}
\begin{center}
\setlength\fboxsep{0pt}
\setlength\fboxrule{0pt}
\framebox[0.15\columnwidth]{\includegraphics[width=0.15\columnwidth]{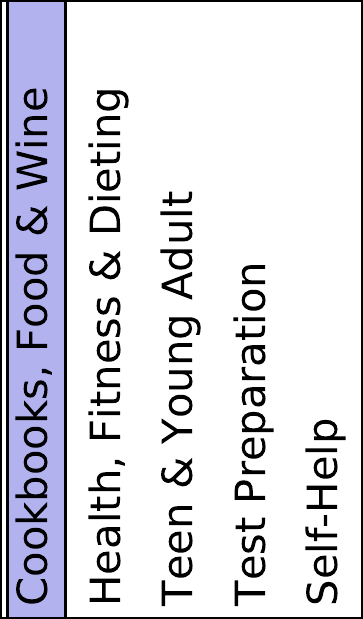}}
\framebox[0.15\columnwidth]{\includegraphics[width=0.15\columnwidth]{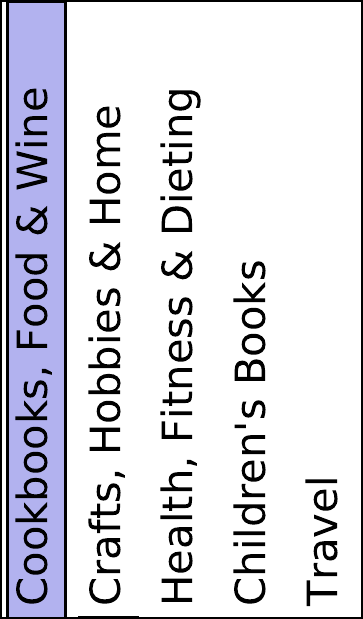}}
\framebox[0.15\columnwidth]{\includegraphics[width=0.15\columnwidth]{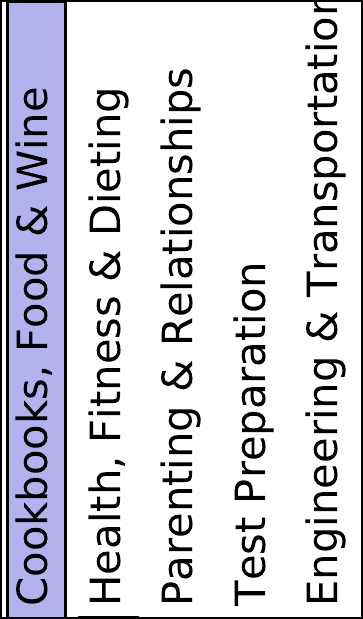}}
\framebox[0.024\columnwidth]{ }
\framebox[0.15\columnwidth]{\includegraphics[width=0.15\columnwidth]{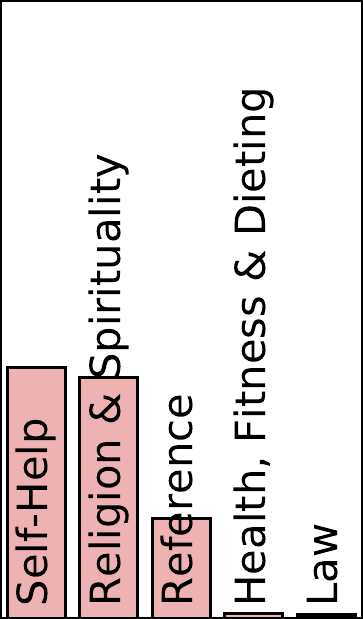}}
\framebox[0.15\columnwidth]{\includegraphics[width=0.15\columnwidth]{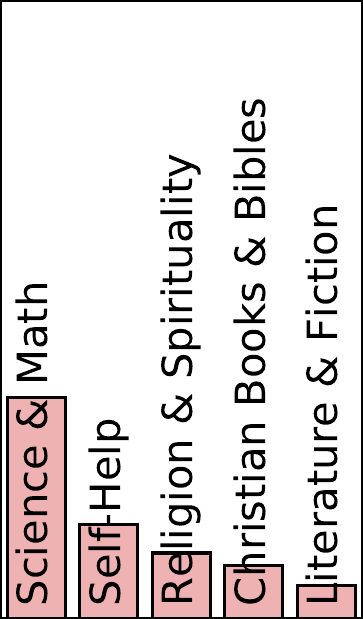}}
\framebox[0.15\columnwidth]{\includegraphics[width=0.15\columnwidth]{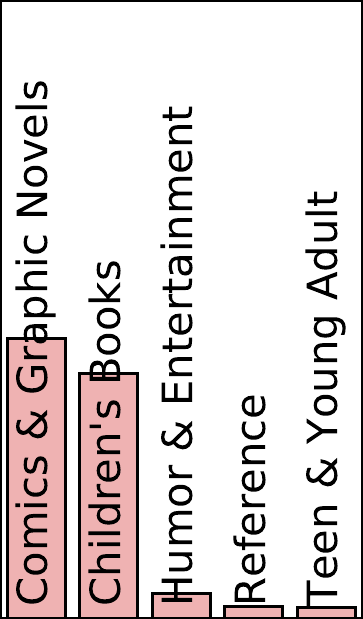}}
\end{center}
\begin{center}
\setlength\fboxsep{1pt}
\setlength\fboxrule{0pt}
\framebox[0.45\columnwidth]{(a)}
\framebox[0.024\columnwidth]{ }
\framebox[0.45\columnwidth]{(b)}
\end{center}
\caption{\label{food} Sample test set images from the ``Cookbooks, Food \& Wine'' category. The top row shows the cover images and the bottom row shows their respective softmax activations from AlexNet. The blue bar is the correct class and the red bars are the other classes. Only the top 5 highest activations are displayed. (a) is examples of correctly classified books and (b) is examples of books belonging to ``Cookbooks, Food \& Wine'' that were misclassified as other classes.  }
\end{figure}

In general, most cover images have either a strong activation toward a single class or are ambiguous and could be part of many classes at once. 
Figure~\ref{food} shows examples of books classified in the ``Cookbooks, Food \& Wine'' category. 
When the cover contained an image of food, the CNN predicted the correct class and with a high probability.
But, the covers with more ambiguous images resulted in a low confidence.
The misclassified examples in Fig.~\ref{food}~(b) failed for understandable reasons;
the first two are ambiguous and can reasonably be classified as ``Self-Help'' and ``Science \& Math'' respectively. 
The final example had a strong probability of being in ``Comics \& Graphic Novels'' and ``Children's Books'' because the cover image features an illustration of a vehicle.
Many books contain misleading covers like these examples and correct classification would be difficult even for a human without reading the text.

\begin{figure}
\begin{center}
\setlength\fboxsep{0pt}
\setlength\fboxrule{0pt}
\framebox[0.13\columnwidth]{\includegraphics[width=0.13\columnwidth]{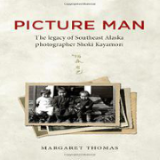}}
\framebox[0.13\columnwidth]{\includegraphics[width=0.13\columnwidth]{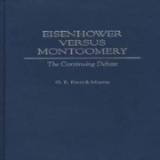}}
\framebox[0.13\columnwidth]{\includegraphics[width=0.13\columnwidth]{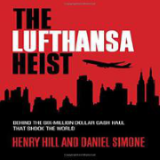}}
\framebox[0.13\columnwidth]{\includegraphics[width=0.13\columnwidth]{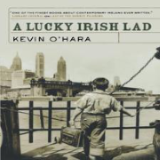}}
\framebox[0.13\columnwidth]{\includegraphics[width=0.13\columnwidth]{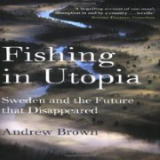}}
\framebox[0.13\columnwidth]{\includegraphics[width=0.13\columnwidth]{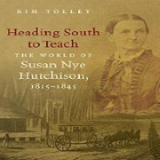}}
\framebox[0.13\columnwidth]{\includegraphics[width=0.13\columnwidth]{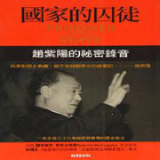}}
\end{center}
\begin{center}
\setlength\fboxsep{0pt}
\setlength\fboxrule{0pt}
\framebox[0.13\columnwidth]{\includegraphics[width=0.13\columnwidth]{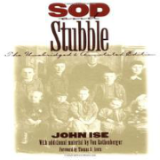}}
\framebox[0.13\columnwidth]{\includegraphics[width=0.13\columnwidth]{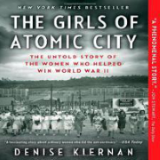}}
\framebox[0.13\columnwidth]{\includegraphics[width=0.13\columnwidth]{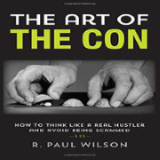}}
\framebox[0.13\columnwidth]{\includegraphics[width=0.13\columnwidth]{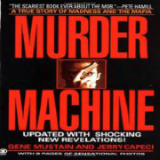}}
\framebox[0.13\columnwidth]{\includegraphics[width=0.13\columnwidth]{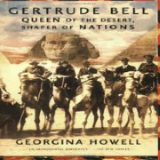}}
\framebox[0.13\columnwidth]{\includegraphics[width=0.13\columnwidth]{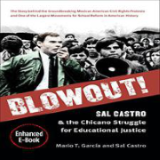}}
\end{center}
\caption{\label{misclassbio} The ``Biographies \& Memoirs'' book covers that were classified by AlexNet as ``History.'' While misclassified, many of these books also can relate to ``History'' despite the ground truth. }
\end{figure}

Figure~\ref{misclassbio} reveals another example of misleading cover images, but for the ``Biographies \& Memoirs'' category.
The difficulty of this category comes from a high rate of sharing qualities with other categories causing substantial ambiguity of the genre itself.
A high number of misclassifications from the ``Biographies \& Memoirs'' category went into ``History.''
However, Fig.~\ref{misclassbio} shows that most of those misclassifications could be considered to be part of both categories.
We also observed a similar relationship between ``Comics \& Graphic Novels'' and ``Children's Books'' and between ``Medical Books'' and ``Science \& Math.''
This shows that the AlexNet network was able to automatically learn relationships between categories based solely on the cover images.

\begin{figure*}[tb]
\begin{center}
\includegraphics[clip,width=1.83\columnwidth,trim={0cm 1.6cm 0cm 1.7cm}]{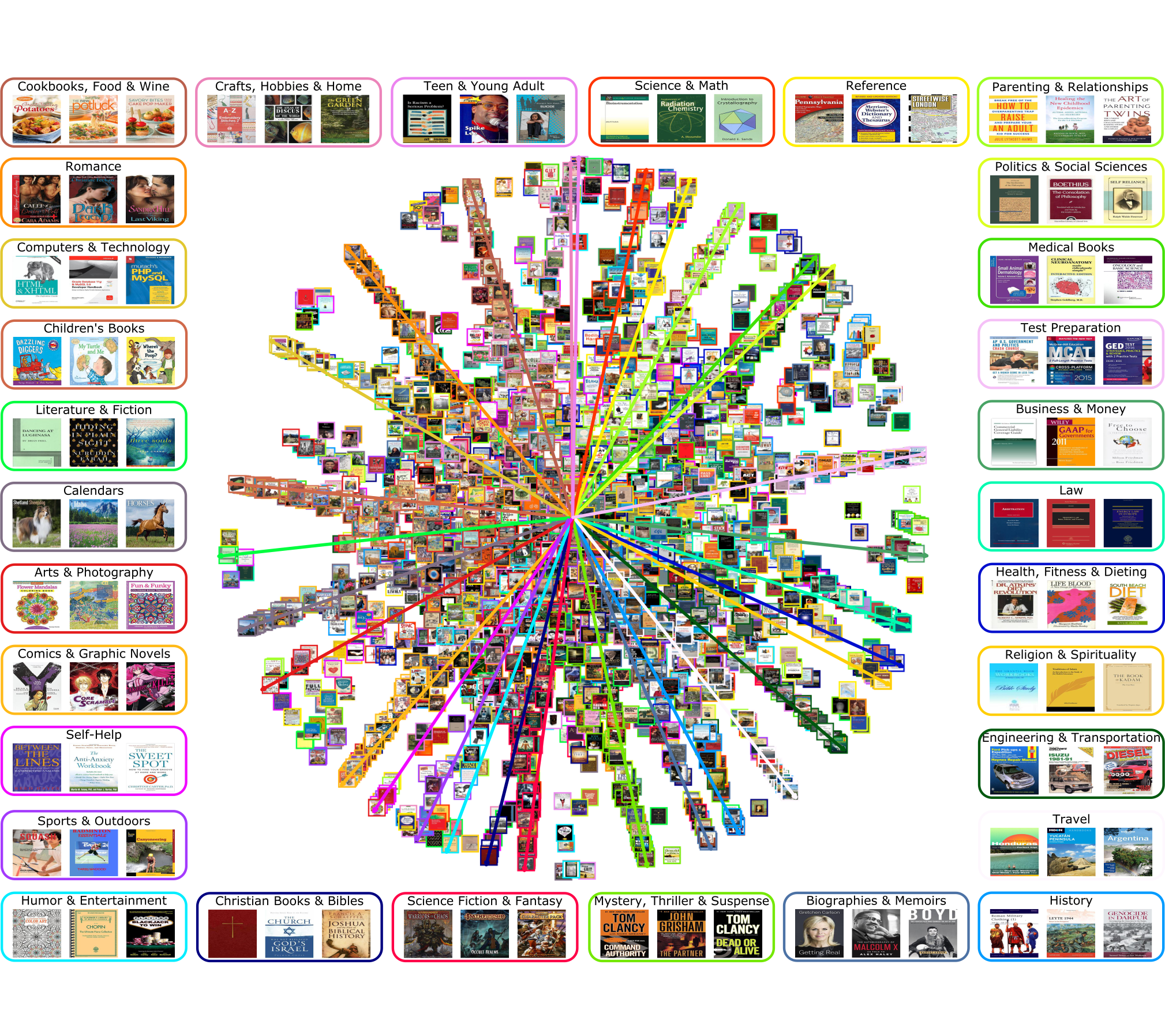} 
\end{center}
\caption{\label{mds} Visualization of the output layer softmax activations of AlexNet. Each point is a 30-dimensional vector where each dimension is the probability of each output class. For visualization purposes, the points are mapped into 2-dimensional subspace with PCA. The arrows represent the axes of each class. The class ground truth is represented by colors, chosen at random. Sample images with high activations from each class are enlarged.}
\end{figure*}

From visualizing the softmax activations in Fig.~\ref{mds}, we can see an overview of the probability of class membership as determined by the network for each of the book covers.
The figure clearly shows the large central cluster of difficult covers as well as the confident correctly classified covers near each axis.
For classes such as ``Politics \& Social Sciences'' and ``Christian Books \& Bibles,'' the strong softmax responses are sparse and it is reflected in their very low recognition accuracy.
Conversely, the densely activated axes have high recognition accuracies indicating that they have unique visual relationships to their genre.

\section{Book Cover Design Principles}
\label{design}

Analysis of the results reveals that AlexNet was able to learn certain high-level features of each category. 
Some of these correlated features may be objects such as portraits for ``Biographies \& Memoirs'' or food for ``Cookbooks, Food \& Wine.''
Other times it is colors, layout, or text.
In this section, we explore the design principles that the CNN was able to automatically learn. 

\subsection{Color Matters}

\begin{figure}
\small
\begin{center}
\setlength\fboxsep{0pt}
\setlength\fboxrule{0pt}
\framebox[0.11\columnwidth]{\includegraphics[width=0.11\columnwidth]{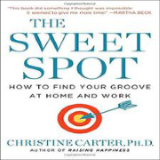}}
\framebox[0.11\columnwidth]{\includegraphics[width=0.11\columnwidth]{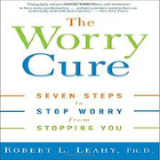}}
\framebox[0.11\columnwidth]{\includegraphics[width=0.11\columnwidth]{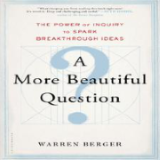}}
\framebox[0.11\columnwidth]{\includegraphics[width=0.11\columnwidth]{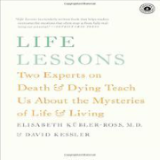}}
\framebox[0.02\columnwidth]{ }
\framebox[0.11\columnwidth]{\includegraphics[width=0.11\columnwidth]{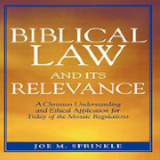}}
\framebox[0.11\columnwidth]{\includegraphics[width=0.11\columnwidth]{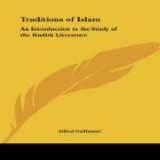}}
\framebox[0.11\columnwidth]{\includegraphics[width=0.11\columnwidth]{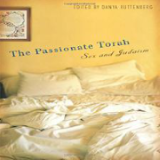}}
\framebox[0.11\columnwidth]{\includegraphics[width=0.11\columnwidth]{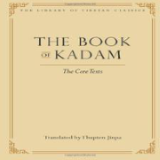}}

\framebox[0.44\columnwidth]{``Self-Help''}
\framebox[0.02\columnwidth]{ }
\framebox[0.44\columnwidth]{``Religion \& Spirituality''}
\framebox[0.44\columnwidth]{White}
\framebox[0.02\columnwidth]{ }
\framebox[0.44\columnwidth]{Yellow}
\end{center}
\begin{center}
\setlength\fboxsep{0pt}
\setlength\fboxrule{0pt}
\framebox[0.11\columnwidth]{\includegraphics[width=0.11\columnwidth]{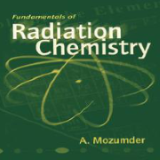}}
\framebox[0.11\columnwidth]{\includegraphics[width=0.11\columnwidth]{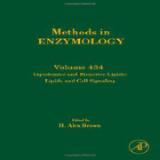}}
\framebox[0.11\columnwidth]{\includegraphics[width=0.11\columnwidth]{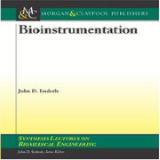}}
\framebox[0.11\columnwidth]{\includegraphics[width=0.11\columnwidth]{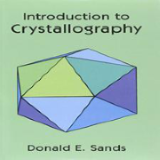}}
\framebox[0.02\columnwidth]{ }
\framebox[0.11\columnwidth]{\includegraphics[width=0.11\columnwidth]{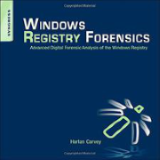}}
\framebox[0.11\columnwidth]{\includegraphics[width=0.11\columnwidth]{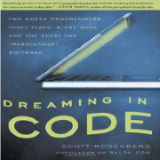}}
\framebox[0.11\columnwidth]{\includegraphics[width=0.11\columnwidth]{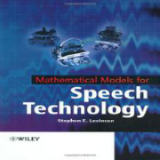}}
\framebox[0.11\columnwidth]{\includegraphics[width=0.11\columnwidth]{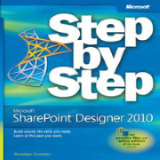}}

\framebox[0.44\columnwidth]{``Science \& Math''}
\framebox[0.02\columnwidth]{ }
\framebox[0.44\columnwidth]{``Computers \& Technology''}
\framebox[0.44\columnwidth]{Green}
\framebox[0.02\columnwidth]{ }
\framebox[0.44\columnwidth]{Blue}
\end{center}
\begin{center}
\setlength\fboxsep{0pt}
\setlength\fboxrule{0pt}
\framebox[0.11\columnwidth]{\includegraphics[width=0.11\columnwidth]{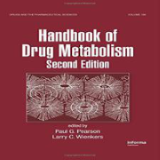}}
\framebox[0.11\columnwidth]{\includegraphics[width=0.11\columnwidth]{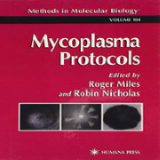}}
\framebox[0.11\columnwidth]{\includegraphics[width=0.11\columnwidth]{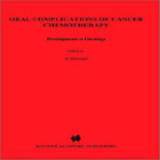}}
\framebox[0.11\columnwidth]{\includegraphics[width=0.11\columnwidth]{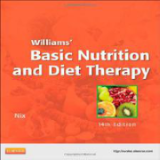}}
\framebox[0.02\columnwidth]{ }
\framebox[0.11\columnwidth]{\includegraphics[width=0.11\columnwidth]{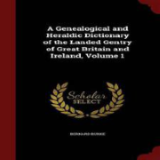}}
\framebox[0.11\columnwidth]{\includegraphics[width=0.11\columnwidth]{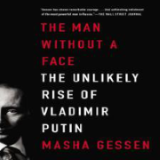}}
\framebox[0.11\columnwidth]{\includegraphics[width=0.11\columnwidth]{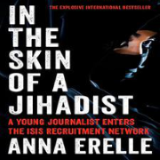}}
\framebox[0.11\columnwidth]{\includegraphics[width=0.11\columnwidth]{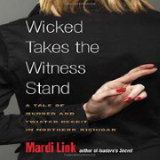}}

\framebox[0.44\columnwidth]{``Medical Books''}
\framebox[0.02\columnwidth]{ }
\framebox[0.44\columnwidth]{``Biographies \& Memoirs''}
\framebox[0.44\columnwidth]{Red}
\framebox[0.02\columnwidth]{ }
\framebox[0.44\columnwidth]{Black}
\end{center}
\caption{\label{misclasspoly} Book covers from genres with particular color associations. Each example was correctly classified by the AlexNet.}
\end{figure}

In the absence of distinguishable features, the CNN has to rely on color alone to classify covers.
Because of this, many classes get associated to certain colors for books with limited features. 
Shown in Fig.~\ref{misclasspoly}, the AlexNet relates white to ``Self-Help,'' yellow to ``Religion \& Spirituality,'' green to ``Science \& Math,'' blue to ``Computers \& Technology,'' red to ``Medical Books,'' and black to ``Biographies \& Memoirs.''
Although, classifying simple book covers by color alone causes many misclassifications to occur.

\begin{figure}
\small
\begin{center}
\setlength\fboxsep{0pt}
\setlength\fboxrule{0pt}
\framebox[0.11\columnwidth]{\includegraphics[width=0.11\columnwidth]{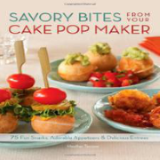}}
\framebox[0.11\columnwidth]{\includegraphics[width=0.11\columnwidth]{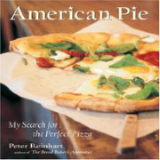}}
\framebox[0.11\columnwidth]{\includegraphics[width=0.11\columnwidth]{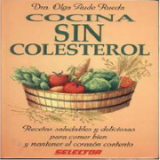}}
\framebox[0.11\columnwidth]{\includegraphics[width=0.11\columnwidth]{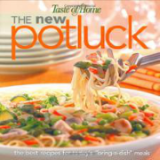}}
\framebox[0.02\columnwidth]{ }
\framebox[0.11\columnwidth]{\includegraphics[width=0.11\columnwidth]{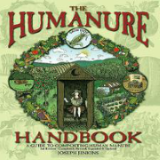}}
\framebox[0.11\columnwidth]{\includegraphics[width=0.11\columnwidth]{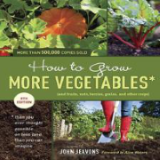}}
\framebox[0.11\columnwidth]{\includegraphics[width=0.11\columnwidth]{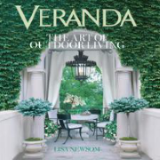}}
\framebox[0.11\columnwidth]{\includegraphics[width=0.11\columnwidth]{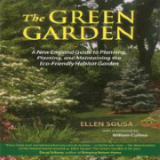}}

\framebox[0.44\columnwidth]{``Cookbooks, Food \& Wine''}
\framebox[0.02\columnwidth]{ }
\framebox[0.44\columnwidth]{``Crafts, Hobbies \& Home''}
\framebox[0.44\columnwidth]{Beige}
\framebox[0.02\columnwidth]{ }
\framebox[0.44\columnwidth]{Green}
\end{center}
\begin{center}
\setlength\fboxsep{0pt}
\setlength\fboxrule{0pt}
\framebox[0.11\columnwidth]{\includegraphics[width=0.11\columnwidth]{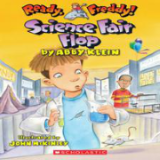}}
\framebox[0.11\columnwidth]{\includegraphics[width=0.11\columnwidth]{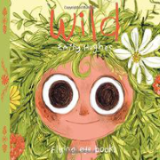}}
\framebox[0.11\columnwidth]{\includegraphics[width=0.11\columnwidth]{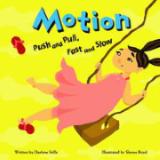}}
\framebox[0.11\columnwidth]{\includegraphics[width=0.11\columnwidth]{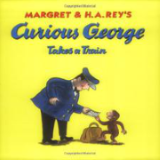}}
\framebox[0.02\columnwidth]{ }
\framebox[0.11\columnwidth]{\includegraphics[width=0.11\columnwidth]{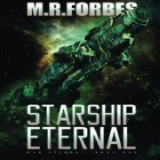}}
\framebox[0.11\columnwidth]{\includegraphics[width=0.11\columnwidth]{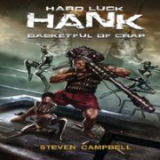}}
\framebox[0.11\columnwidth]{\includegraphics[width=0.11\columnwidth]{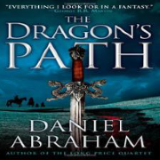}}
\framebox[0.11\columnwidth]{\includegraphics[width=0.11\columnwidth]{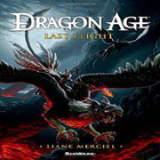}}

\framebox[0.44\columnwidth]{``Children's Books''}
\framebox[0.02\columnwidth]{ }
\framebox[0.44\columnwidth]{``Science Fiction \& Fantasy''}
\framebox[0.44\columnwidth]{Bright}
\framebox[0.02\columnwidth]{ }
\framebox[0.44\columnwidth]{Dark}
\end{center}
\caption{\label{mood} Book covers that were successfully classified by the common moods or color pallets of respective genres.}
\end{figure}

The color association does not only restrict itself to simple book covers. 
Despite having active book covers, the tone of book covers were also important for classification.
For example, ``Cookbooks, Food \& Wine'' often features food and are commonly by shades of beige and tan (Fig.~\ref{mood}).
Likewise, there is a high representation of gardening books in the ``Crafts, Hobbies \& Home'' class, therefore, green books are commonly classified in that genre.
Also, the tone of the book can define the mood, so ``Children's Books'' commonly have designs with yellow or bright backgrounds and ``Science Fiction \& Fantasy'' books usually have black or dark backgrounds.
The AlexNet was able to successfully capture the mood of book genres by grouping books of certain moods to respective genres.

\subsection{Objects Matter}
\begin{figure}
\small
\begin{center}
\setlength\fboxsep{2pt}
\setlength\fboxrule{0pt}
\framebox[0.11\columnwidth]{\includegraphics[width=0.11\columnwidth]{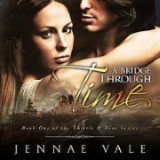}}
\framebox[0.11\columnwidth]{\includegraphics[width=0.11\columnwidth]{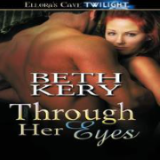}}
\framebox[0.11\columnwidth]{\includegraphics[width=0.11\columnwidth]{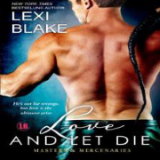}}
\framebox[0.11\columnwidth]{\includegraphics[width=0.11\columnwidth]{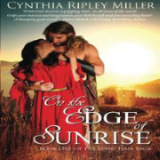}}
\framebox[0.02\columnwidth]{ }
\framebox[0.11\columnwidth]{\includegraphics[width=0.11\columnwidth]{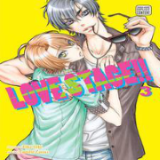}}
\framebox[0.11\columnwidth]{\includegraphics[width=0.11\columnwidth]{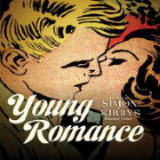}}
\framebox[0.11\columnwidth]{\includegraphics[width=0.11\columnwidth]{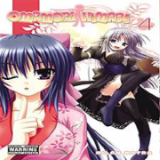}}
\framebox[0.11\columnwidth]{\includegraphics[width=0.11\columnwidth]{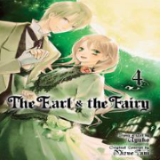}}

\framebox[0.44\columnwidth]{``Romance''}
\framebox[0.02\columnwidth]{ }
\framebox[0.44\columnwidth]{``Comics \& Graphic Novels''}
\framebox[0.44\columnwidth]{Intimate}
\framebox[0.02\columnwidth]{ }
\framebox[0.44\columnwidth]{Illustrated}
\end{center}
\begin{center}
\setlength\fboxsep{0pt}
\setlength\fboxrule{0pt}
\framebox[0.11\columnwidth]{\includegraphics[width=0.11\columnwidth]{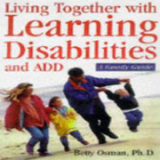}}
\framebox[0.11\columnwidth]{\includegraphics[width=0.11\columnwidth]{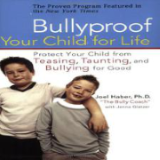}}
\framebox[0.11\columnwidth]{\includegraphics[width=0.11\columnwidth]{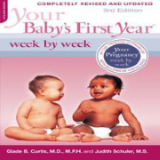}}
\framebox[0.11\columnwidth]{\includegraphics[width=0.11\columnwidth]{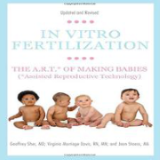}}
\framebox[0.02\columnwidth]{ }
\framebox[0.11\columnwidth]{\includegraphics[width=0.11\columnwidth]{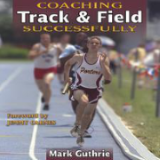}}
\framebox[0.11\columnwidth]{\includegraphics[width=0.11\columnwidth]{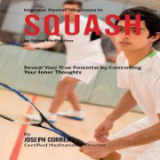}}
\framebox[0.11\columnwidth]{\includegraphics[width=0.11\columnwidth]{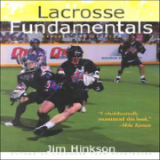}}
\framebox[0.11\columnwidth]{\includegraphics[width=0.11\columnwidth]{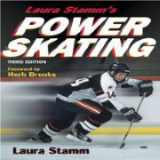}}

\framebox[0.44\columnwidth]{``Parenting \& Relationships''}
\framebox[0.02\columnwidth]{ }
\framebox[0.44\columnwidth]{``Sports \& Outdoors''}
\framebox[0.44\columnwidth]{Young}
\framebox[0.02\columnwidth]{ }
\framebox[0.44\columnwidth]{Active}
\end{center}
\begin{center}
\setlength\fboxsep{0pt}
\setlength\fboxrule{0pt}
\framebox[0.11\columnwidth]{\includegraphics[width=0.11\columnwidth]{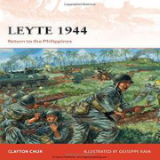}}
\framebox[0.11\columnwidth]{\includegraphics[width=0.11\columnwidth]{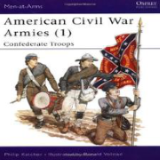}}
\framebox[0.11\columnwidth]{\includegraphics[width=0.11\columnwidth]{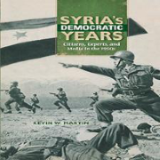}}
\framebox[0.11\columnwidth]{\includegraphics[width=0.11\columnwidth]{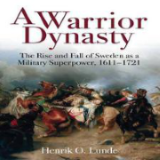}}
\framebox[0.02\columnwidth]{ }
\framebox[0.11\columnwidth]{\includegraphics[width=0.11\columnwidth]{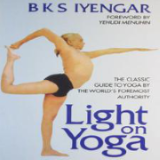}}
\framebox[0.11\columnwidth]{\includegraphics[width=0.11\columnwidth]{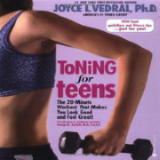}}
\framebox[0.11\columnwidth]{\includegraphics[width=0.11\columnwidth]{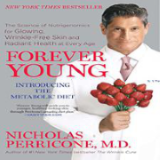}}
\framebox[0.11\columnwidth]{\includegraphics[width=0.11\columnwidth]{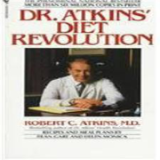}}

\framebox[0.44\columnwidth]{``History''}
\framebox[0.02\columnwidth]{ }
\framebox[0.44\columnwidth]{``Health, Fitness \& Dieting''}
\framebox[0.44\columnwidth]{Soldiers}
\framebox[0.02\columnwidth]{ }
\framebox[0.44\columnwidth]{Exercise or Doctors}
\end{center}
\caption{\label{people} Correctly classified book covers that feature different aspects of humans.}
\end{figure}

The image on book covers is usually the thing that first attracts potential readers to a book.
It should be no surprise that the object featured on the cover has an effect on how it gets classified.
What is surprising about the results of our experiment is how the network is able to distinguish different genres but with common objects.
For instance, featuring people on the cover is common among many genres, but the type of person or how the person is dressed determines how the book gets classified.
Figure~\ref{people} shows four genres that centrally display humans, but have discriminating features that make the classes separable.

\begin{figure}
\small
\begin{center}
\setlength\fboxsep{0pt}
\setlength\fboxrule{0pt}
\framebox[0.11\columnwidth]{\includegraphics[width=0.11\columnwidth]{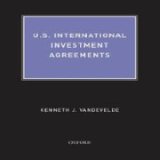}}
\framebox[0.11\columnwidth]{\includegraphics[width=0.11\columnwidth]{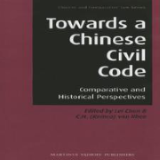}}
\framebox[0.11\columnwidth]{\includegraphics[width=0.11\columnwidth]{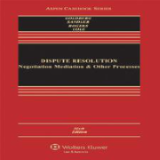}}
\framebox[0.11\columnwidth]{\includegraphics[width=0.11\columnwidth]{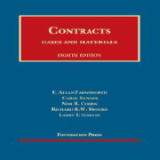}}
\framebox[0.02\columnwidth]{ }
\framebox[0.11\columnwidth]{\includegraphics[width=0.11\columnwidth]{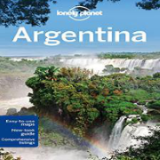}}
\framebox[0.11\columnwidth]{\includegraphics[width=0.11\columnwidth]{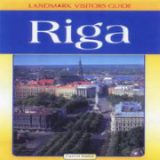}}
\framebox[0.11\columnwidth]{\includegraphics[width=0.11\columnwidth]{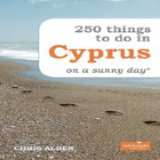}}
\framebox[0.11\columnwidth]{\includegraphics[width=0.11\columnwidth]{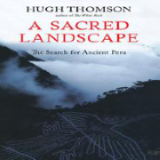}}

\framebox[0.44\columnwidth]{``Law''}
\framebox[0.02\columnwidth]{ }
\framebox[0.44\columnwidth]{``Travel''}
\framebox[0.44\columnwidth]{Title Boards}
\framebox[0.02\columnwidth]{ }
\framebox[0.44\columnwidth]{Landscape Photographs}
\end{center}
\caption{\label{layout} Examples of layout considerations as determined by the AlexNet for ``Law'' and ``Travel.''.}
\end{figure}

The structure and layout of the book cover also makes a difference in the classification.
Books with rectangular title boards, no matter the color, tended to be classified as ``Law'' and books with a large landscape photographs tended to be ``Travel'' (Fig.~\ref{layout}).
This trend continued to other categories, such as ``Cookbooks, Food \& Wine'' with a central image of food stretching to the edges of the cover, ``Biographies \& Memoirs'' featuring close-up shots of people, and reference and textbooks containing solid color bands.

\subsection{Text Matters}
\begin{figure}
\small
\begin{center}
\setlength\fboxsep{2pt}
\setlength\fboxrule{0pt}
\framebox[0.11\columnwidth]{\includegraphics[width=0.11\columnwidth]{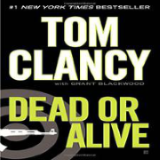}}
\framebox[0.11\columnwidth]{\includegraphics[width=0.11\columnwidth]{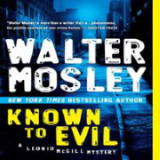}}
\framebox[0.11\columnwidth]{\includegraphics[width=0.11\columnwidth]{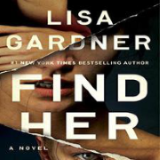}}
\framebox[0.11\columnwidth]{\includegraphics[width=0.11\columnwidth]{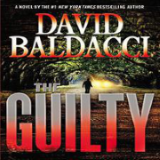}}
\framebox[0.02\columnwidth]{ }
\framebox[0.11\columnwidth]{\includegraphics[width=0.11\columnwidth]{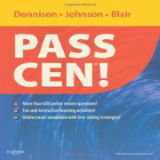}}
\framebox[0.11\columnwidth]{\includegraphics[width=0.11\columnwidth]{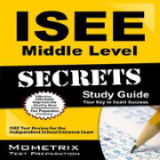}}
\framebox[0.11\columnwidth]{\includegraphics[width=0.11\columnwidth]{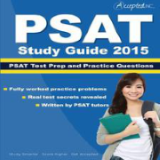}}
\framebox[0.11\columnwidth]{\includegraphics[width=0.11\columnwidth]{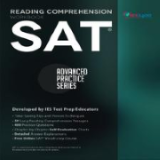}}

\framebox[0.44\columnwidth]{``Mystery, Thriller \& Suspense''}
\framebox[0.02\columnwidth]{ }
\framebox[0.44\columnwidth]{``Test Preparation''}
\framebox[0.44\columnwidth]{Large Overlaid Text}
\framebox[0.02\columnwidth]{ }
\framebox[0.44\columnwidth]{Large but Short Text}
\end{center}
\begin{center}
\setlength\fboxsep{0pt}
\setlength\fboxrule{0pt}
\framebox[0.11\columnwidth]{\includegraphics[width=0.11\columnwidth]{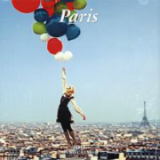}}
\framebox[0.11\columnwidth]{\includegraphics[width=0.11\columnwidth]{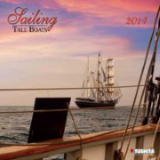}}
\framebox[0.11\columnwidth]{\includegraphics[width=0.11\columnwidth]{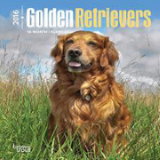}}
\framebox[0.11\columnwidth]{\includegraphics[width=0.11\columnwidth]{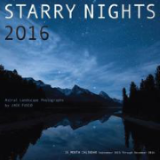}}
\framebox[0.02\columnwidth]{ }
\framebox[0.11\columnwidth]{\includegraphics[width=0.11\columnwidth]{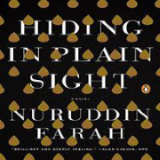}}
\framebox[0.11\columnwidth]{\includegraphics[width=0.11\columnwidth]{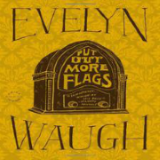}}
\framebox[0.11\columnwidth]{\includegraphics[width=0.11\columnwidth]{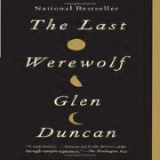}}
\framebox[0.11\columnwidth]{\includegraphics[width=0.11\columnwidth]{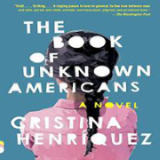}}

\framebox[0.44\columnwidth]{``Calendars''}
\framebox[0.02\columnwidth]{ }
\framebox[0.44\columnwidth]{``Literature \& Fiction''}
\framebox[0.44\columnwidth]{Sparse Text}
\framebox[0.02\columnwidth]{ }
\framebox[0.44\columnwidth]{Expressive Fonts}
\end{center}
\caption{\label{font} Book covers showing text and font differences.}
\end{figure}

Another interesting design principle captured by the AlexNet is the text qualities and the font properties.
The best example of this is ``Mystery, Thriller \& Suspense,'' shown in Fig.~\ref{font}.
Despite having a similar color pallet and image content to ``Romance'' and ``Science Fiction \& Fantasy,'' the common thread in many of the classified ``Mystery, Thriller \& Suspense'' books was large overlaid sans serif text. 
Figure~\ref{font} also shows that ``Calendars'' often de-emphasize the title text so the focus is on the cover image.
On the other hand, the figure also shows that ``Literature \& Fiction'' often uses expressive fonts to reveal messages about the book.
The text style on the cover of a book affects the classification, revealing that relationships between text style and genre exist.

\begin{figure}
\begin{center}
\setlength\fboxsep{1pt}
\setlength\fboxrule{0pt}
\framebox[0.145\columnwidth]{\includegraphics[width=0.14\columnwidth]{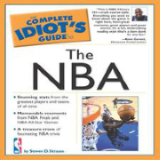}}
\framebox[0.145\columnwidth]{\includegraphics[width=0.14\columnwidth]{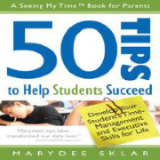}}
\framebox[0.145\columnwidth]{\includegraphics[width=0.14\columnwidth]{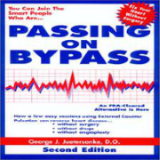}}
\framebox[0.145\columnwidth]{\includegraphics[width=0.14\columnwidth]{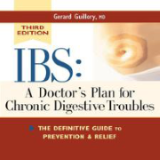}}
\framebox[0.145\columnwidth]{\includegraphics[width=0.14\columnwidth]{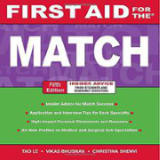}}
\framebox[0.145\columnwidth]{\includegraphics[width=0.14\columnwidth]{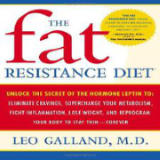}}

\end{center}
\caption{\label{misclasstest} Books from other categories that were classified as ``Test Preparation.'' The correct labels for the books from left to right are ``Sports \& Outdoors,'' ``Parenting \& Relationships,'' ``Medical Books,'' ``Health, Fitness \& Dieting,'' ``Health, Fitness \& Dieting,'' and ``Cookbooks, Food \& Wine.'' }
\end{figure}

In particular, of the 30 classes, ``Test Preparation'' had the highest recognition rate at 68.9\%, much higher than the overall accuracy.
The reason behind this high accuracy is that ``Test Preparation'' book covers are often formulaic. 
They tend to have an acronym in large letters (e.g.  ``SAT,'' ``GRE,'' ``GMAT,'' etc.) near the top with horizontal or vertical stripes and possibly a small image of people.
The large text is important because when compared to other non-fiction and reference classes, the presence of large acronyms is the most discriminating factor.
Figure~\ref{misclasstest} shows books from other categories that were incorrectly classified as ``Test Preparation.''
These examples follow the design rules similar to many other ``Test Preparation'' books, but the actual content of the text reveals the books as other classes.

\section{Conclusion}
\label{conclusion}
In this paper, we presented the application of machine learning to predict the genre of a book based on its cover image.
We showed that it is possible to draw a relationship between book cover images and genre using automatic recognition.
Using a CNN model, we categorized book covers into genres and the results of using AlexNet with transfer learning had an accuracy of 24.7\% for Top~1, 33.1\% for Top~2, and 40.3\% for Top~3 in 30-class classification.
The 5-layer LeNet had a lower accuracy of 13.5.7\% for Top~1, 21.4\% for Top~2, and 27.8\% for Top~3.
Using the pre-trained AlexNet had a dramatic effect on the accuracy compared to the LeNet.

However, classification of books based on the cover image is a difficult task.
We revealed that many books have cover images with few visual features or ambiguous features causing for many incorrect predictions.
While uncovering some of the design rules found by the CNN, we found that books can have also misleading covers.
In addition, because books can be part of multiple genres, the CNN had a poor Top~1 performance.
To overcome this, experiments can be done using multi-label classification.

Future research will be put into further analysis of the characteristics of the classifications and the features determined by the network in an attempt to design a network that is optimized for this task.
Increasing the size of the network or tuning the hyperparameters may improve the performance.
In addition, the book cover dataset we created can be used for other tasks as it contains other information such as title, author, and category hierarchy.
Genre classification can also be done using supplemental information such as textual features alongside the cover images.
We hope to design more robust models to better capture the essence of cover design.

\section*{Acknowledgments}
This research was partially supported by MEXT-Japan (Grant No. 26240024) and the Institute of Decision Science for a Sustainable Society, Kyushu University, Fukuoka, Japan.

All book cover images are copyright Amazon.com, Inc. The display of the images are transformative and are used as fair use for academic purposes.

The book cover database is available at \url{https://github.com/uchidalab/book-dataset}. 

\bibliographystyle{IEEEtran}
\bibliography{deep,master,books}



%

\end{document}